\documentclass[letterpaper, 10 pt, conference]{ieeeconf}  

\IEEEoverridecommandlockouts                              

\usepackage{graphics} 
\usepackage{bm}
\usepackage{float} 
\usepackage{stfloats}
\usepackage{caption}
\usepackage{cancel}
\usepackage{epsfig} 
\usepackage{overpic}
\usepackage{times} 
\usepackage{booktabs}
\usepackage{rotating}
\usepackage{algorithmic}
\usepackage{algorithm}

\usepackage{amsmath} 
\usepackage{amssymb}  

\usepackage{amsthm}

\usepackage{mathrsfs}
\usepackage[table, dvipsnames, svgnames, x11names]{xcolor} 
\usepackage{colortbl}
\usepackage{multirow}
\usepackage{marvosym}
\definecolor{mygray}{gray}{.9}
\usepackage{pifont}
\usepackage{cite}
\let\labelindent\relax
\usepackage{enumitem}
\usepackage[colorlinks,
            linkcolor=blue,
            anchorcolor=blue,
            citecolor=green,
            urlcolor=violet,
            backref=page]{hyperref}

\title{\LARGE \bf
GAT-Grasp: Gesture-Driven Affordance Transfer for Task-Aware \\Robotic Grasping}

\author{Ruixiang Wang$^{1, 2*}$, Huayi Zhou$^{1*}$, Xinyue Yao$^{1,3}$, Guiliang Liu$^{1}$, Kui Jia$^{1\dagger}$
\thanks{*These authors contributed equally to this work.} 
\thanks{$^{\dagger}$Corresponding author:  kuijia@gmail.com.}%
\thanks{$^{1}$The Chinese University of Hong Kong, Shenzhen.}
\thanks{$^{2}$Harbin Institute of Technology.}
\thanks{$^{3}$Zhejiang University.}
}

\begin{document}

\makeatletter
\let\@oldmaketitle\@maketitle
\renewcommand{\@maketitle}{\@oldmaketitle
\vspace{0.4cm}
\centering
  \includegraphics[width=0.98\linewidth]
    {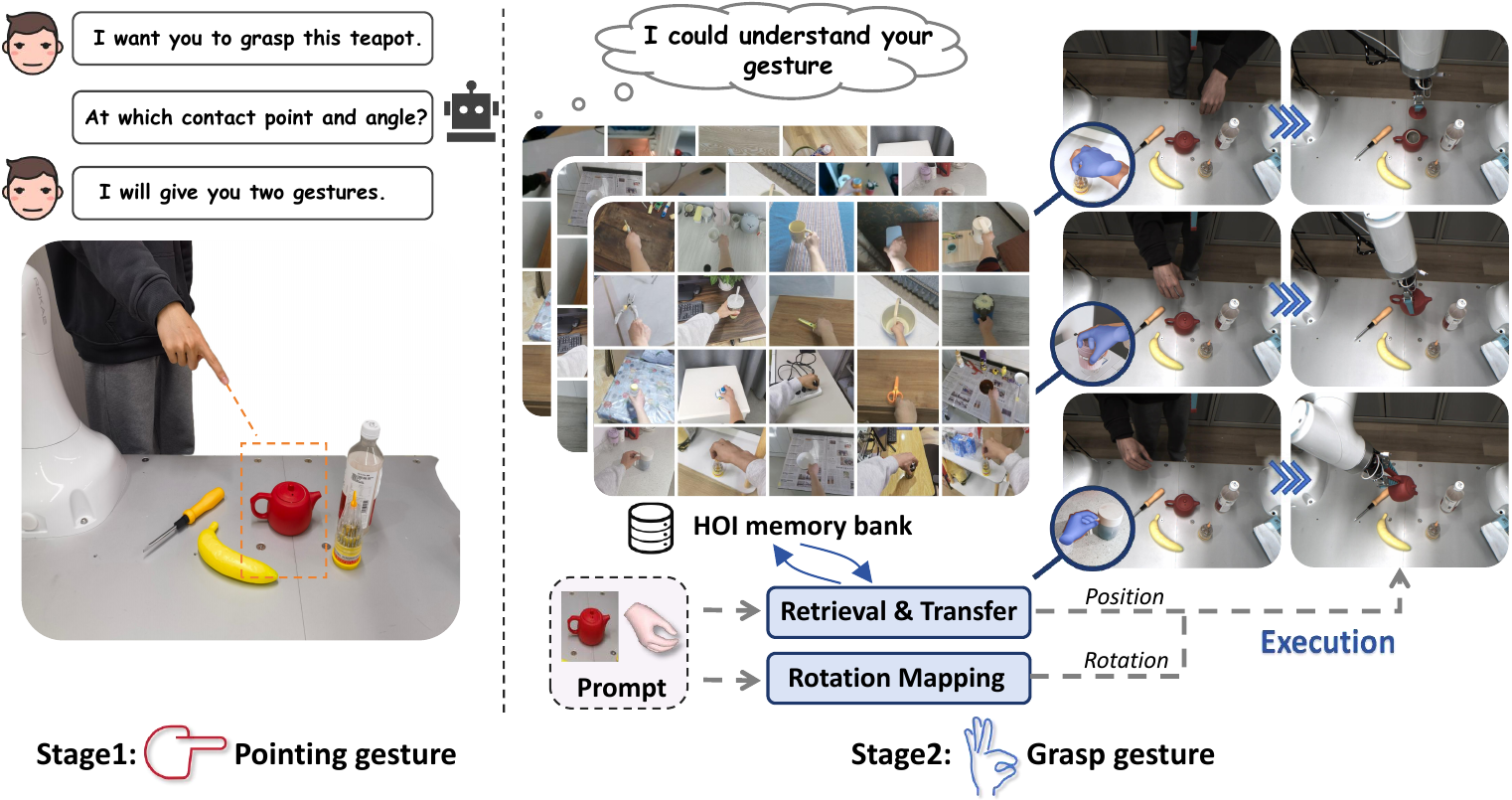}
    \captionof{figure}{Overview of \textbf{GAT-Grasp} (\textbf{G}esture-Driven \textbf{A}ffordance \textbf{T}ransfer for Task-Aware Robotic Grasping). A pointing gesture (left) specifies the target region, while a grasp gesture (right) encodes the intended grasping strategy. Through retrieval-based affordance transfer and rotation mapping, GAT-Grasp accurately interprets human intent and executes precise, task-specific grasps on the robot.}
    \label{fig: main}\vspace{-0.1cm}
    }%
\makeatother

\maketitle
\thispagestyle{empty}
\pagestyle{empty}
\addtocounter{figure}{-1}

\begin{abstract}

Achieving precise and generalizable grasping across diverse objects and environments is essential for intelligent and collaborative robotic systems. However, existing approaches often struggle with ambiguous affordance reasoning and limited adaptability to unseen objects, leading to suboptimal grasp execution. In this work, we propose GAT-Grasp, a gesture-driven grasping framework that directly utilizes human hand gestures to guide the generation of task-specific grasp poses with appropriate positioning and orientation. Specifically, we introduce a retrieval-based affordance transfer paradigm, leveraging the implicit correlation between hand gestures and object affordances to extract grasping knowledge from large-scale human-object interaction videos. By eliminating the reliance on pre-given object priors, GAT-Grasp enables zero-shot generalization to novel objects and cluttered environments. Real-world evaluations confirm its robustness across diverse and unseen scenarios, demonstrating reliable grasp execution in complex task settings.

\end{abstract}

\section{INTRODUCTION}

Seamless collaboration between humans and robots requires robotic systems to accurately perceive, interpret, and respond to human intent under a variety of environments. To achieve this, various communication modalities (such as language instructions\cite{kuangRAMRetrievalBasedAffordance2024a, maGLOVERGeneralizableOpenVocabulary2024}, eye gazing\cite{9395580}, click-based input on images\cite{tsagkasClickGraspZeroShot2024,nohGraspSAMWhenSegment2024}, and hand gesture\cite{constantin2022interactive,medeiros20213d}) have been widely explored in human-robot interaction (HRI). These modalities serve as natural interfaces, helping robots understand human intent and make informed decisions when interacting with objects. While significant progress has been made in object-level manipulation, achieving precise partial object grasping remains a fundamental challenge, as it demands a fine-grained spatial understanding of affordance regions on an object.

Most existing approaches rely on language-based guidance to instruct robotic grasping and interaction\cite{brohan2023can,luVLGrasp6DofInteractive2023,zhengGaussianGrasper3DLanguage2024a}. However, language descriptions often remain to be coarse-grained without specifying precise grasping locations. For instance, while directing a robot to "pick up a bottle" is straightforward, specifying "grasp the bottle cap to unscrew it" requires a more location-informed spatial reference that is difficult to convey through natural language. This lack of precision in spatial referencing not only limits the robot's ability to perform fine-grained tasks but also hinders its generalization to unseen objects and scenarios, where task-specific affordances are critical.

Humans naturally adapt their grasp strategies based on an object's shape, function, and intended use, encoding task-relevant affordances in their actions. For example, as illustrated in Fig. \ref{fig: main}, grasping a teapot handle to lift it differs significantly from grasping its lid to open it. These grasp variations inherently reflect affordance cues and can serve as structured guidance for robotic grasping. However, transferring such human grasp strategies to robots requires a systematic approach to affordance learning, transfer, and execution.

To address this challenge, we propose \textbf{GAT-Grasp}, a gesture-conditioned grasping pipeline that enables robots to infer both "where" and "how" to grasp in a task-specific manner. The pipeline begins with a pointing gesture, which provides an initial localization of the target graspable region. A subsequent grasp gesture refines the robot's understanding of the precise grasp location and orientation. To support this process, we construct an affordance memory based on Hand-Object Interaction (HOI) data\cite{damenScalingEgocentricVision2018,grauman2022ego4d,liuHOI4D4DEgocentric2022}, serving as a structured knowledge base for grasp affordance transfer. Using a hierarchical retrieval strategy, the system searches for semantically and visually similar instances, enabling the transfer of grasp strategies to unseen target objects. Furthermore, we develop a hand-gesture-to-gripper mapping module, which extracts the rotational properties of human grasps and translates them into the robot’s end-effector constraints, ensuring precise grasp execution, which is the crucial aspect for task-specific grasping and tool-use scenarios. Finally, we integrate the retrieved affordance regions and rotation constraints into off-the-shelf grasp generators, producing task-conditioned grasp poses that align with human intent.

We conduct extensive real-world evaluations to validate our method, demonstrating significant improvements over existing task-oriented grasping techniques. In particular, our method outperforms the state-of-the-art in challenging cluttered object scenarios. These results highlight the effectiveness of affordance-driven grasping, paving the way for more intuitive and task-aware robotic manipulation. Our work makes the following key contributions:
\begin{itemize}
\item We introduce a novel hand-conditioned grasping pipeline that takes RGB-D scene inputs and grasp gestures to generate precise, task-aware grasp poses, demonstrating superior performance over language-based approaches.

\item We propose a retrieval-based affordance transfer mechanism that maps grasp gestures to task-specific affordance regions, enabling more precise and intent-aligned grasp localization than pointing alone.

\item We develop a hand-gripper mapping module that converts human grasp intent into gripper-compatible orientations, ensuring adaptive, task-specific execution.


\end{itemize}

\section{Related Work}

\subsection{Human-Robot Interaction}

Human-robot interaction (HRI) is fundamental for seamless collaboration, enabling effective intent communication and shared autonomy across diverse applications\cite{goodrichHumanRobotInteraction2008,bonarini2020communication}. Advances in natural language processing allow robots to process spoken and written instructions, providing a flexible and high-level communication channel for robotic control\cite{lu2023vl,tziafasOpenWorldGraspingLarge2024a}. Beyond language, vision-based modalities like gesture inputs offer more intuitive and spatially efficient communication. Recent works have explored pointing gestures for guiding robotic actions\cite{jojic2000detection,dhingra2020recognition,das2021data,azari2019commodifying}. However, pointing alone suffers from ambiguities in cluttered environments, where multiple objects may align with the pointing direction, leading to grasping uncertainties\cite{das2021data}. To address this, our pipeline integrates both pointing and grasp gestures, capturing the geometric and semantic properties of objects to infer accurate graspable regions and orientations.

\subsection{Task-Oriented Grasping}

Task-oriented grasping aims to select grasp configurations that align with specific object affordances and facilitate downstream tasks such as pouring, cutting, and assembling. Recent approaches leverage the reasoning capabilities of LLMs and vision-language models (VLMs) to infer task-relevant object parts for grasping\cite{huang2023voxposer,qianThinkGraspVisionLanguageSystem2024,maGLOVERGeneralizableOpenVocabulary2024}. For example, ThinkGrasp~\cite{qianThinkGraspVisionLanguageSystem2024} combines LLM-based contextual reasoning with vision models like SAM~\cite{kirillov2023segment} and VLPart~\cite{sunGoingDenserOpenVocabulary2023a} to segment affordance regions, while RoboPoint~\cite{yuanRoboPointVisionLanguageModel2024} fine-tunes VLMs for open-vocabulary affordance reasoning. However, these methods often struggle with ambiguous affordance cues and depend heavily on visual model performance. Eye-gaze-based approaches use human gaze patterns to identify task-relevant areas\cite{8793804,ryu2019gg,wang2023you}, but they require specialized hardware and are sensitive to calibration errors. Click-based methods specify grasp points in 2D images\cite{tsagkasClickGraspZeroShot2024, nohGraspSAMWhenSegment2024}, but their reliance on computer interfaces limits real-world applicability. In contrast, our method directly conveys grasping intent through hand gestures, bypassing the need for precise visual grounding and enabling direct affordance localization, offering a more intuitive, robust, and annotation-efficient solution for task-aware robotic grasping.

\subsection{Affordance Learning from Human Hand Video}
The vast availability of human-object interaction videos on the internet provides a rich resource for learning affordances and manipulation skills from human demonstrations\cite{grauman2022ego4d,liuHOI4D4DEgocentric2022,damenScalingEgocentricVision2018}. During these interactions, human hands naturally reveal key affordance cues, such as contact points, hand poses, and motion trajectories, which can be leveraged for robotic learning. Recent works have explored imitation learning from human demonstration videos\cite{li2024okami,zhouYouOnlyTeach2025}, but these approaches often struggle with generalization beyond controlled environments. Robo-ABC\cite{juRoboABCAffordanceGeneralization2024} and RAM\cite{kuangRAMRetrievalBasedAffordance2024a} address this by constructing an affordance memory bank, using CLIP\cite{radford2021learning} to retrieve similar demonstrations for unseen objects. Inspired by their approach, we also adopt a retrieval-based learning strategy. However, our method incorporates hand gestures as retrieval queries, enabling affordance learning from the implicit interaction between human hands and object geometry. This allows our approach to generalize affordance reasoning to unseen objects using unlabeled human demonstration videos, eliminating the need for explicit language annotations and enhancing scalability in real-world scenarios.

\section{Method}

 This section introduces the implementation details of GAT-Grasp. Given a visual observation \( O \) from a binocular camera, we aim to grasp a specific part using two hand gestures. The first Pointing gesture \( P \) specifies the approximate region of the target grasping area (\S\ref{sec:pointing_gesture}), while the second Grasp gesture \( G \) learns the more precise position (\S\ref{sec:retrival}) and angle (\S\ref{sec:mapping}), ultimately generating a robust grasp (\S\ref{sec:pose}).

\subsection{Pointing Gesture for Target Region Localization} 
\label{sec:pointing_gesture}
The pointing gesture serves as the initial step to localize the target region for grasping. Given a visual observation \( O \), the hand gesture is first processed using WiLoR~\cite{potamiasWiLoREndtoend3D2024} to generate key points \( \{ k_{p,j} \}_{j=1}^n \). Specifically, the key points of the index finger \( \ k_{\text{index}} \) are used to define a pointing direction. Traditional methods often directly use \( \{ k_{p,j} \}_{j=1}^n \) estimated by the model. However, in practical applications, we observe that 3D shape regression methods such as WiLoR~\cite{potamiasWiLoREndtoend3D2024} and HaMeR\cite{pavlakos2024reconstructing} exhibit non-negligible errors, leading to inaccuracies in the generated 3D key points. To address this issue, we propose an alternative approach that leverages 2D-to-3D projection and stereo matching for improved accuracy.

First, all 3D key points \( \{ k_{p,j} \}_{j=1}^n \) are projected onto the 2D image plane. 
Next, the depth of each key point is computed using a stereo matching algorithm~\cite{xu2023iterative}, resulting in a set of refined 3D points \( \{ \widehat{k}_{p,j} \}_{j=1}^n \), where each \( \widehat{k}_{p,j} = (u_j, v_j, d_j) \) represents the 3D coordinates of the \( j \)-th index finger key point in the camera frame. A 3D line \( \mathbf{r}(t) = \mathbf{o} + t \mathbf{d} \) is fitted to the inlier points, where \( \mathbf{o} \in \mathbb{R}^3 \) is the origin (typically the wrist joint), \( \mathbf{d} \in \mathbb{R}^3 \) is the unit direction vector of the ray, and \( t \) is a scalar parameter. To further improve robustness, the RANSAC algorithm\cite{fischler1981random} is applied to remove outliers in the index finger's key points. The intersection of this line with the depth map is computed to determine the target point \( \mathbf{p}^* \in \mathbb{R}^3 \), formulated as:
\begin{equation} 
\mathbf{p}^* = \{ \mathbf{p} \in \mathcal{P} \mid \|\mathbf{r}(t) - \mathbf{p}\| < \epsilon \}  
\end{equation}
where \( \mathcal{P} \) represents the 3D coordinate of the depth map. The 2D coordinates \( \mathbf{u}^* = (u^*, v^*) \) of \( \mathbf{p}^* \) are subsequently extracted. Using \( \mathbf{u}^* \) as the center, a rectangular region of interest \(I^T\) is cropped from the image, defining the approximate target area for grasping.

\label{sec:retrival}

\begin{figure}[t]  
    \centering  
    \includegraphics[width=0.48\textwidth]{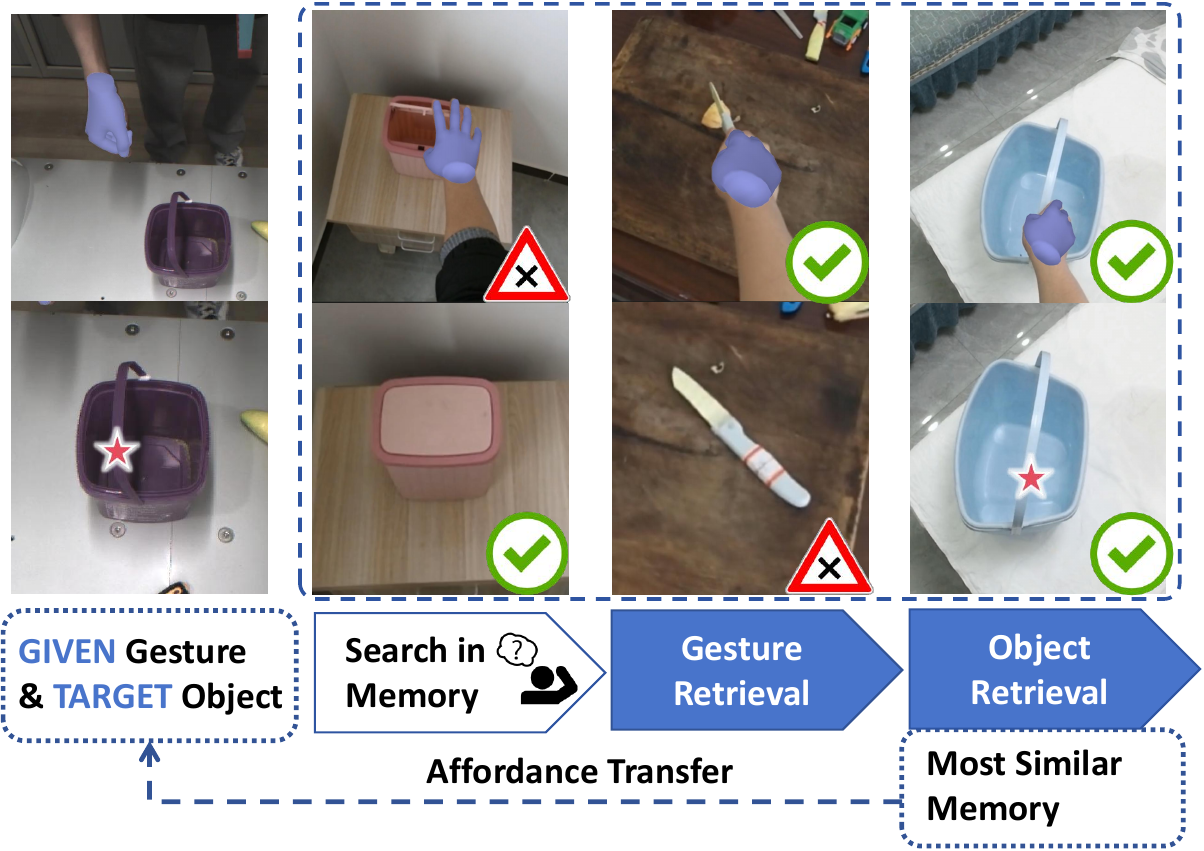}
    \caption{Illustration of our retrieval and transfer pipeline. Given a grasp gesture, our method accurately locates the robot grasp point \textcolor{red}{${\star}$} on the target object.}  
    \label{fig: hand_retrieve} 
\end{figure} 

\subsection{Grasp Gesture Retrieval}

During human-object interaction, rich affordance information is inherently encoded in contact points, hand poses, and grasping strategies. By extracting and modeling this information, we establish a structured mapping between hand gestures and object affordances, enabling robots to interpret human intent and execute precise grasps. However, a key challenge lies in generalizing grasp gestures across diverse objects, as affordance cues vary with object geometry, material, and task-specific constraints. To address this, we construct an affordance memory bank that captures the correspondence between grasp gestures and affordance representations across diverse objects. This allows robots to retrieve and transfer affordance knowledge, facilitating versatile and precise grasping in novel scenarios. The overall process is shown in Fig. \ref{fig: hand_retrieve}.

\textbf{Affordance Memory Bank.}  
To effectively transfer human grasp strategies to robotic manipulation, we construct an Affordance Memory Bank, a structured repository encoding gesture-conditioned affordance priors from large-scale human-object interaction videos. This memory serves as a foundation for retrieving task-relevant grasp affordances, enabling zero-shot generalization to unseen objects. Our memory bank consists of triplets:
\begin{equation}
\mathcal{M} = \{(G_{m,i}^\diamond, I^S_i, C^S_i)\}_{i=1}^{N}
\end{equation}
where \(G_m^\diamond\) represents the grasp gesture extracted using WiLoR, \(I^S\) denotes the source object-centered image, and \(C^S\) specifies the contact point associated with the grasp interaction, implemented similarly to Robo-ABC\cite{juRoboABCAffordanceGeneralization2024}. Each element in the memory bank encodes a real-world grasp interaction.

To achieve generalization across objects, it is crucial to collect a comprehensive set of affordance data covering diverse objects and grasp gestures. The internet provides a rich source of human-object interaction (HOI) videos\cite{damenScalingEgocentricVision2018,grauman2022ego4d,liuHOI4D4DEgocentric2022}, making large-scale egocentric video datasets ideal for affordance learning. We extract a subset of videos from HOI4D\cite{liuHOI4D4DEgocentric2022}, a large-scale HOI dataset, to construct our affordance memory bank. This subset includes 12 object categories suitable for grasping. Additionally, we incorporate manually collected data with 32 grasp poses across 12 object categories, ensuring diverse affordance representation. Our memory bank covers nearly all interaction gestures relevant to object manipulation and captures various geometric structures of object parts, enabling effective cross-object affordance transfer.

\begin{algorithm}[t]
\caption{Hand Gesture Retrieval}
\begin{algorithmic}\small
    \STATE \textbf{Input:} Query grasp gesture $G^{\diamond}_q$, memory bank $\{G^{\diamond}_{m,i}\}_{i=1}^N$, reference keypoints $(k_\text{wri}, k_\text{ind}, k_\text{pin})$, chirality $\diamond = L$ or $\diamond = R$.  

    \STATE \textbf{Step 1: Gesture Alignment}  
    \STATE \textbf{for} each grasp gesture $G^{\diamond}_i \in \{ G_q \} \cup \{ G^{\diamond}_m \}$ \textbf{do}  
    \STATE \quad Extract keypoints: $\mathbf{K_i} = \{ k_{i,j} \}_{j=1}^{n} \leftarrow \textbf{WiLoR}(G^{\diamond}_i)$;  
    \STATE \quad \quad $l_\text{x} \leftarrow (k_{i,\text{ind}} - k_{i,\text{wri}})$; \hfill\textcolor{olive}{// X-axis direction}  
    \STATE \quad \quad $v_z \leftarrow \textsc{cross\_product}(l_\text{x}, (k_{i,\text{pin}} - k_{i,\text{wri}}))$; \hfill\textcolor{olive}{// Z-axis}  
    \STATE \quad \quad $\bar{v}_z \leftarrow v_z / (\textsc{normalize}(v_z) + \text{1e-8})$; \hfill\textcolor{olive}{// Normalized Z-axis}  
    \STATE \quad \quad $\bar{v}_x \leftarrow l_\text{x} / (\textsc{normalize}(l_\text{x)} + \text{1e-8})$; \hfill\textcolor{olive}{// Normalized X-axis}  
    \STATE \quad \quad $\bar{v}_y \leftarrow \textsc{cross\_product}(\bar{v}_z, \bar{v}_x)$; \hfill\textcolor{olive}{// Y-axis direction}  
    \STATE \quad \quad $R_{\text{align}, i} \leftarrow \textsc{concatenate}([\bar{v}_x, \bar{v}_y, \bar{v}_z])$; \hfill\textcolor{olive}{// Rotation matrix}  
    \STATE \quad \quad $\mathbf{\tilde{K}_i} = R_{\text{align}, i} \cdot \mathbf{K_i}$; \hfill\textcolor{olive}{// Apply rotation}  
    \STATE \quad \quad $\mathbf{\tilde{K}_i} \leftarrow \mathbf{\tilde{K}_i} / ||l_\text{x}||$; \hfill\textcolor{olive}{// Scale normalization}  
    \STATE \textbf{end for}  

    \STATE \textbf{Step 2: Top-K Gesture Retrieval}  
    \STATE Compute similarity for all stored gestures $G_m$:  
    \STATE \quad $S(G_q, G_m) = \cos(\mathbf{\tilde{K}_q}, \mathbf{\tilde{K}_m})$; \hfill\textcolor{olive}{// Cosine similarity}  
    \STATE Retrieve Top-$K$ most similar gestures $\{G^{\diamond}_m\}_K$ and corresponding images $\{I^S\}_K$;  
    \STATE \textbf{return} $\{G_{m,i}^\diamond\}_{i=1}^K$, $\{I^S_i\}_{i=1}^K$;
\end{algorithmic}
\label{algHandRetrieve} 
\end{algorithm}

\textbf{Hierarchical Retrieval.}  
To ensure accurate affordance transfer, we employ a two-stage hierarchical retrieval strategy to identify the most relevant memory instance.

In the first stage, we align both the query grasp gesture \( G_q^\diamond \) and the stored grasp gestures $\{G^{\diamond}_{m,i}\}_{i=1}^N$ in the memory bank to a canonical coordinate system, ensuring a consistent comparison across different orientations and scales. To achieve this, we select three key points as anatomical landmarks—the wrist \( k_{\text{wri}} \), the proximal phalanx of the index finger \( k_{\text{ind}} \), and the proximal phalanx of the pinky finger \( k_{\text{pin}} \)—as reference points, as they exhibit minimal relative movement across different hand poses. The complete hand retrieval process is summarized in Alg. \ref{algHandRetrieve}, yielding the Top-K most similar grasp gestures \( \{G_{m,i}^\diamond\}_{i=1}^K \) along with their corresponding images \( \{I^S_i\}_{i=1}^K \) in memory bank.

In the second stage, we perform image retrieval to identify the most contextually relevant object affordance. Given the target image \( I^T \) centered around the interaction point, we retrieve the closest matching source image \( I^S \) by computing the cosine similarity:  

\begin{equation}
S(I^T, I^S_i) = \cos(f_{\text{CLIP}}(I^T), f_{\text{CLIP}}(I^S_i))
\end{equation}
where \( f_{\text{CLIP}}(\cdot) \) represents the CLIP\cite{radford2021learning} image encoder. The object image \( I^S \) and its corresponding contact point \( C^S \), which achieve the highest similarity score, are selected for affordance transfer.

\textbf{Affordance Transfer.}
Leveraging dense feature maps extracted by visual foundation models, we transfer the affordance information from the retrieved object to the target object. Specifically, we extract the diffusion features (DIFT)\cite{tangEmergentCorrespondenceImage2023}, which encode rich pixel-to-pixel correspondences between \( I^S \) and \( I^T \). By aligning their spatial structures, the retrieved contact point \( C^S \) is projected onto the target image, yielding the target contact point \( C^T \). This transferred contact point serves as the reference for final grasp generation.

\subsection{Hand Gesture Mapping to Gripper }
\label{sec:mapping}

The gripper's orientation during grasp execution is crucial for stability and task success. For example, picking up a bowl of food requires a more horizontal angle to avoid spilling its contents. Therefore, we aim to incorporate rotational constraints when providing a grasp gesture, allowing the expected orientation to be conveyed to the robot. 

\begin{figure}[htbp] 
    \centering  
    \includegraphics[width=0.48\textwidth]{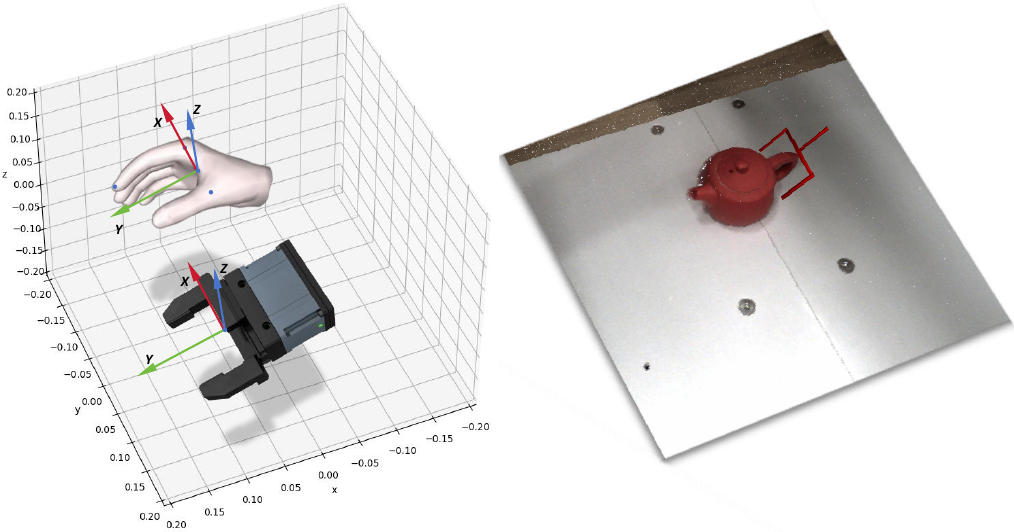}
    \caption{Schematic diagram of hand gesture mapping to the gripper. (Left) Alignment of the gesture coordinate system with the gripper coordinate system. (Right) Demonstration of gripper grasping using the computed rotation angle.}  
    \label{fig: hand_mapping_to_gripper}   
\end{figure}

We introduce a hand gesture mapping module that transforms human grasp gestures into 3D orientations \( R_h \in \mathbb{SO}(3)\) for a two-finger gripper. A gripper-aligned coordinate system is defined using three hand reference points: the proximal phalanx of the thumb \( k_\text{thu} \), the proximal phalanx of the index finger \( k_\text{inp} \), and the fingertip of the index finger \( k_\text{inf} \). The X-axis is the vector from \( k_\text{thu} \) to \( k_\text{inp} \), the XY-plane is formed by the three reference points, and the  Z-axis is the plane's normal, as shown in Fig. \ref{fig: hand_mapping_to_gripper}. Using this formulation, we compute the rotation matrix \( R_h \), which is then used to constrain the final grasp pose.

\subsection{More Robust Grasp Pose Generation}
\label{sec:pose}

With the affordance contact point \( C^T \) and the hand gesture-derived rotation matrix \( R_h \), we can directly generate a 7-DOF grasp poses \( G \), which is defined as $
G = [ t \in \mathbb{R}^{3\times1}, R \in \mathbb{SO}(3), w \in \{0, 1\}]$, where \( t \), \( R \), and \( w \) represent the translation, rotation, and gripper width (open or closed), respectively. This allows our method to operate independently, generating a complete grasp pose without relying on additional grasp generation models. However, to further improve the grasp quality by incorporating geometric information and improving robustness, we optionally integrate state-of-the-art grasp generation models such as Anygrasp et al.~\cite{chen2023efficient, fang2023anygrasp}. 

\begin{figure*}[t]
    \centering  
    \includegraphics[width=1.0\textwidth]{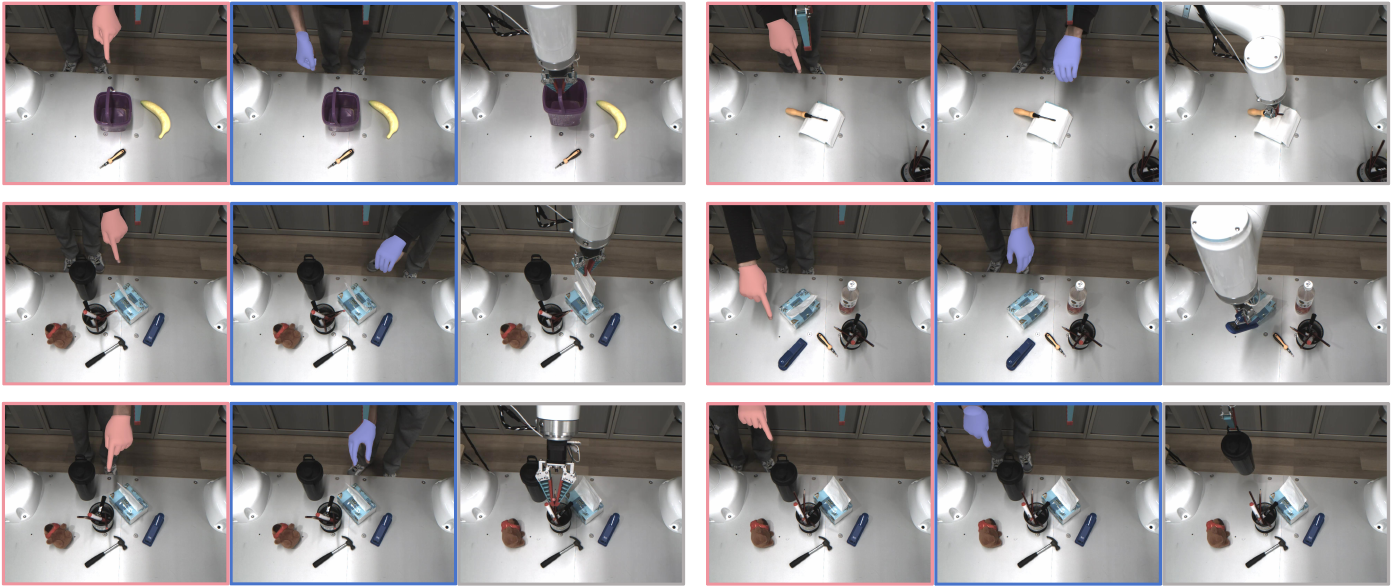}
    \caption{Visualization of different grasping tasks performed on real robots. The pointing gesture is represented in \textcolor{pink}{pink} color, while the grasp gesture is depicted in \textcolor{blue}{blue} color. The robot then executes the corresponding actions based on these gestures.}  
    \label{fig: exp} 
\end{figure*} 

In our implementation, we use HGGD\cite{chen2023efficient} for its scalability. When HGGD is employed, we apply a Gaussian attention mechanism centered at \( C^T \) to refine grasp candidates. This mechanism filters the image to focus on high-affordance regions, ensuring grasp poses are localized near \( C^T \). The resulting grasp candidates can be expressed as:

\begin{equation}
\{(G_i, s_i)\}_{i=1}^{N} = f_{\text{grasp}}(C^T, O)
\end{equation}
where \( O \) represents the observed scene, \( G_i \) is the generated grasp pose, and \( s_i \) is the predicted grasp confidence score.  

To ensure grasp stability and alignment with human intent, we select the optimal grasp \( G^* \) by maximizing both the confidence score \( s_i \) and the alignment between the generated rotation matrix \( R_i \) and the desired rotation \( R_h \). The optimal grasp is given by: 

\begin{equation}
G^* = \arg\max_{G_i} \left( s_i - \lambda \| I - R_h^T R_i \|_F \right)
\end{equation}
where \( \| I - R_h^T R_i \|_F \) measures orientation deviation, and \( \lambda \) balances grasp confidence and orientation alignment.

This formulation ensures the selected grasp has a high success probability and adheres to the intended orientation, improving reliability and task performance. By making HGGD optional, our method offers flexibility: it can directly generate grasp poses using only \( C^T \) and \( R_h \), or leverage the grasp generation model for enhanced geometric reasoning and robustness.

\section{Experiments}

We demonstrate the effectiveness of our pipeline in extensive real-world experiments. In this section, we aim to answer the following questions while presenting a comprehensive evaluation of GAT-Grasp: 1) Can hand gestures serve as a simple and effective way for humans to interact with robots in real-world scenarios? 2) Does GAT-Grasp outperform existing methods in accurate affordances reasoning? 3) Can our approach generate more stable grasp poses, leading to higher grasp success rates?

\subsection{Experimental Setup}

Our system is built on an Rokae SR3 robotic arm\footnote{\url{https://www.rokae.com/cn/product/show/1662.html?ly=serch}}, a 6-DoF manipulator with a span of approximately 705 mm, equipped with a WHEELTEC two-finger gripper\footnote{\url{https://www.wheeltec.net/product/html/?159.html}}  
(maximum opening width 120 mm). For perception, we use the DexSense 3D industrial camera\footnote{\url{https://dexforce-3dvision.com/productinfo/1022811.html}}), which captures stereo image pairs and generates RGB-D images using a stereo matching algorithm \cite{xu2023iterative}. The camera is statically mounted to capture both scene context and hand gestures. All algorithms are deployed on a workstation equipped with an Intel Core i7-13700K CPU and an NVIDIA RTX 4060 GPU.

To evaluate GAT-Grasp, we conduct experiments in two settings: single-object grasping and cluttered-scene grasping. In the single-object setting, we test objects with distinct affordance regions, randomly placed within the robot's workspace. The cluttered scene setting involves multiple objects in unstructured arrangements, challenging the system to resolve affordance ambiguities and occlusions.  Our real-world grasping deployment is shown in Fig. \ref{fig: exp}.

\begin{table*}[t]
    \centering
    \renewcommand{\arraystretch}{1.2} 
    \setlength{\tabcolsep}{6pt}  
    \caption{Quantitative analysis of grasping experiment results in cluttered scenes. 20 grasps are performed on each object part. \textbf{Gesture} refers to the grasp gesture used as a guide input in our method.}
    \label{tab1}
    \begin{tabular}{l ccc ccc cc cc c}
        \toprule
        \multirow{2}{*}{Object Part} & bottle & bottle & bottle & pen & pen & \multirow{2}{*}{stapler} 
        & hammer & hammer & tissue &\multirow{2}{*}{\textbf{AVG}} \\

        & handle & body & cap & holder & in holder & & head & handle & paper  & \\
        \midrule
        \raisebox{3.0\height}{Gesture} & \includegraphics[width=0.04\textwidth]{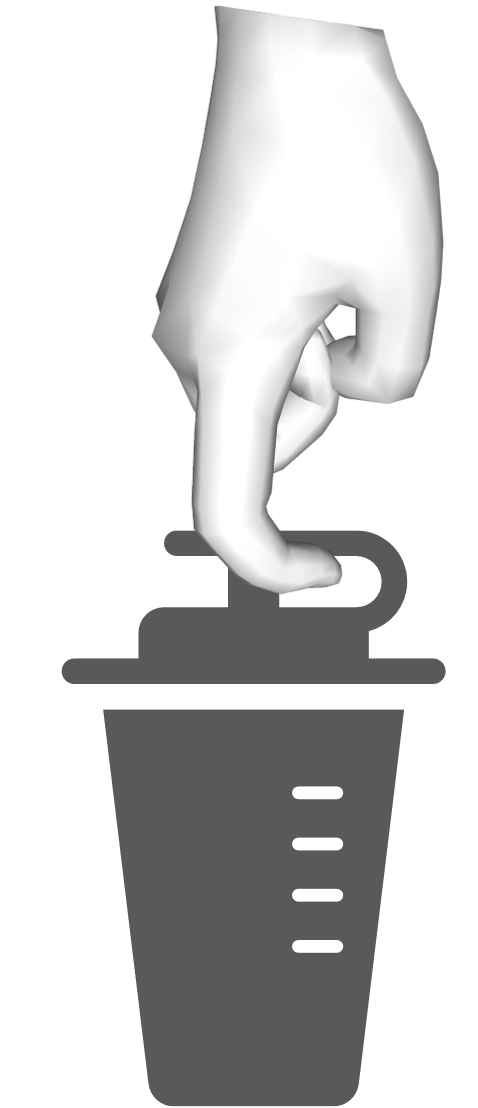} & \includegraphics[width=0.06\textwidth]{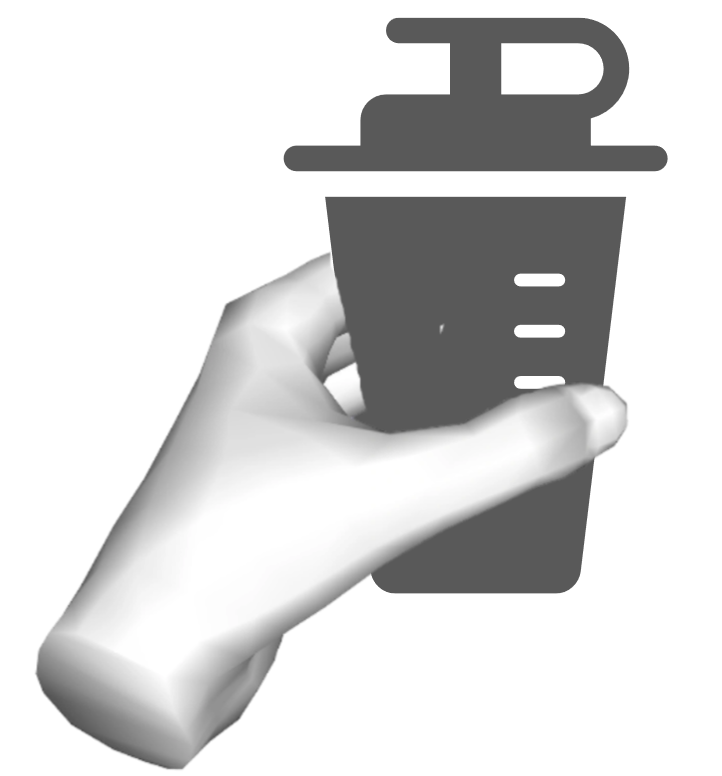}&
        \includegraphics[width=0.05\textwidth]{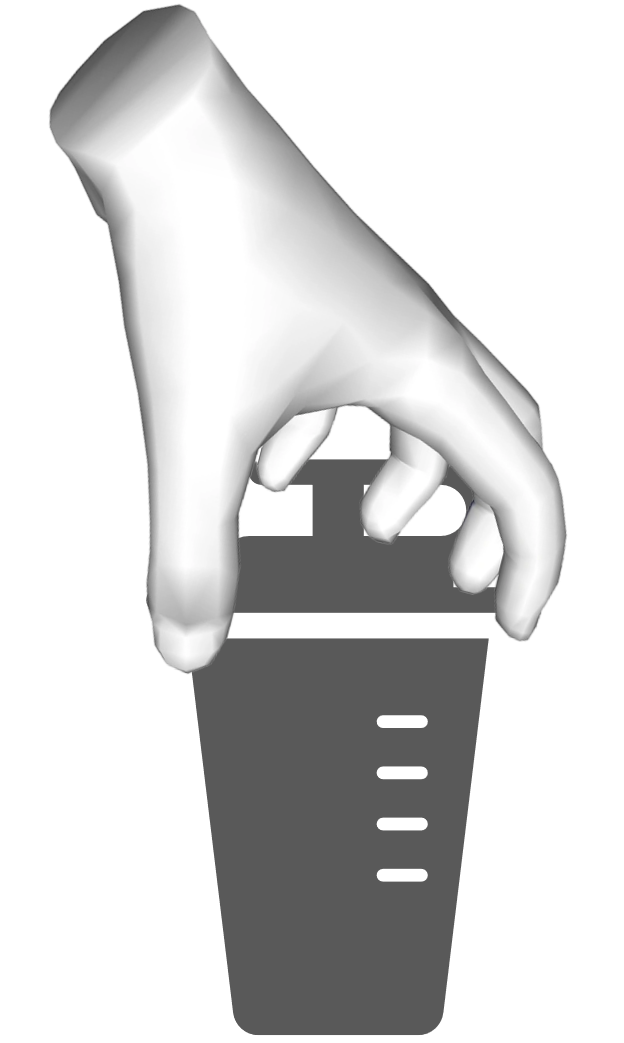}  & \includegraphics[width=0.07\textwidth]{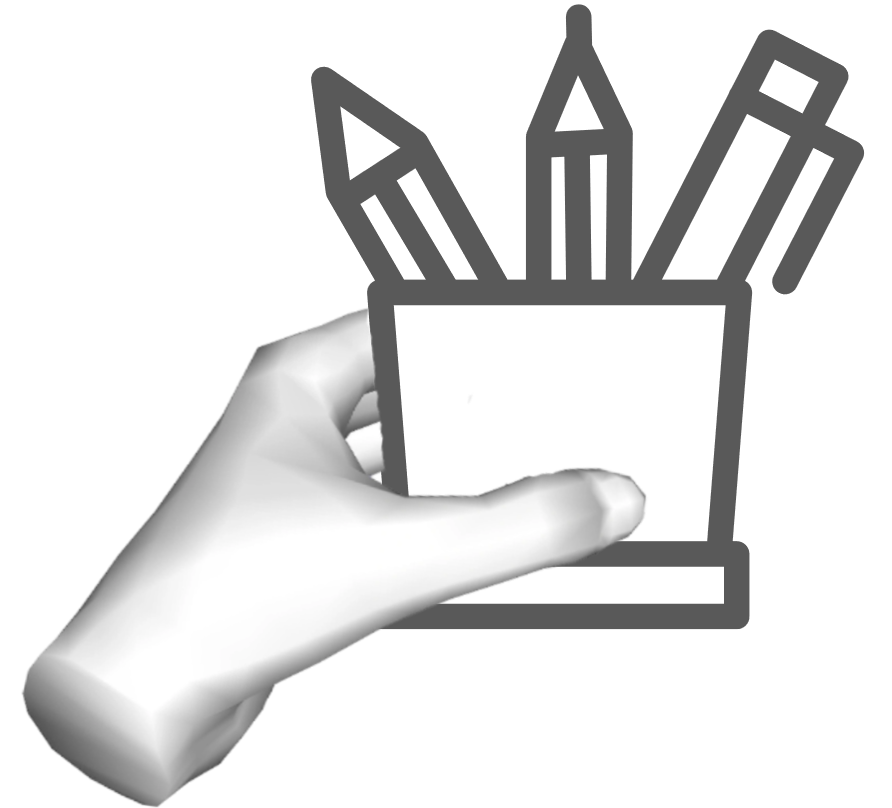}  & \includegraphics[width=0.06\textwidth]{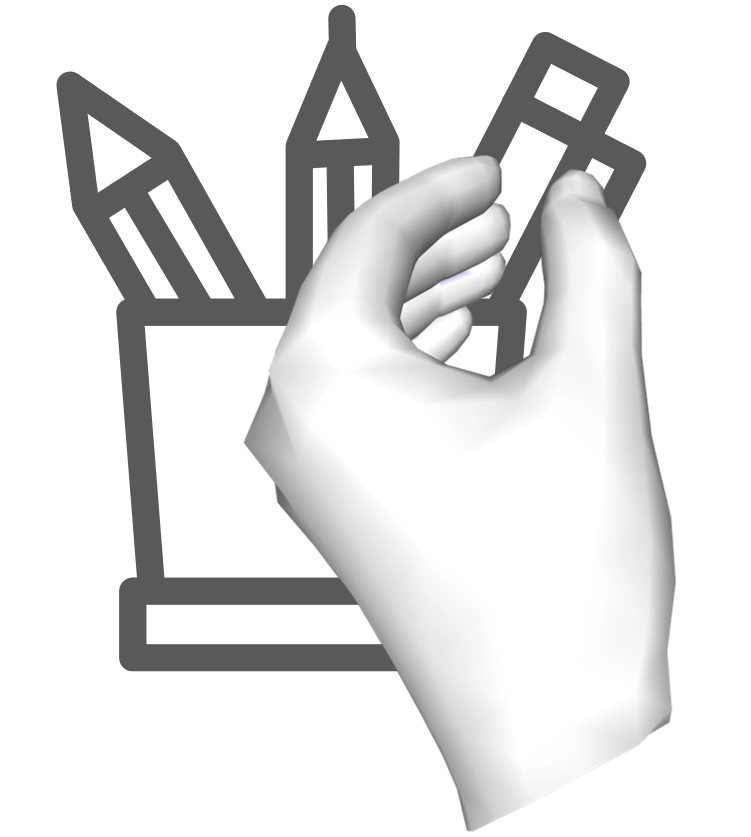} & \includegraphics[width=0.06\textwidth]{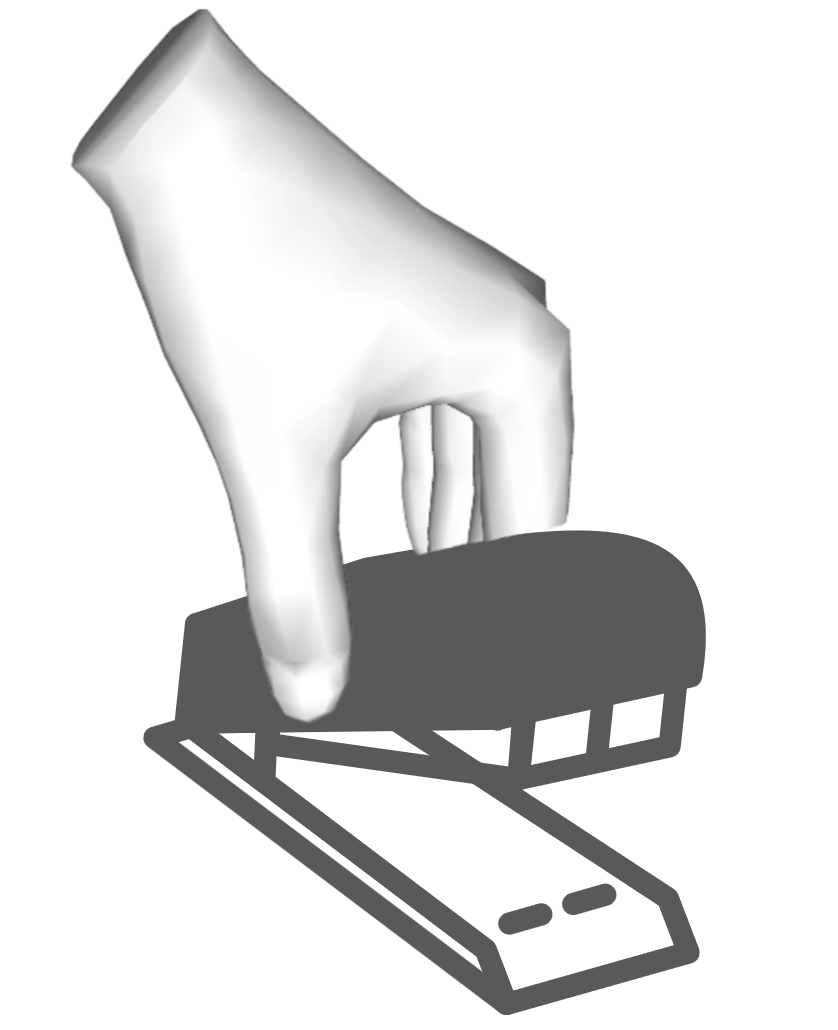}  & \includegraphics[width=0.06\textwidth]{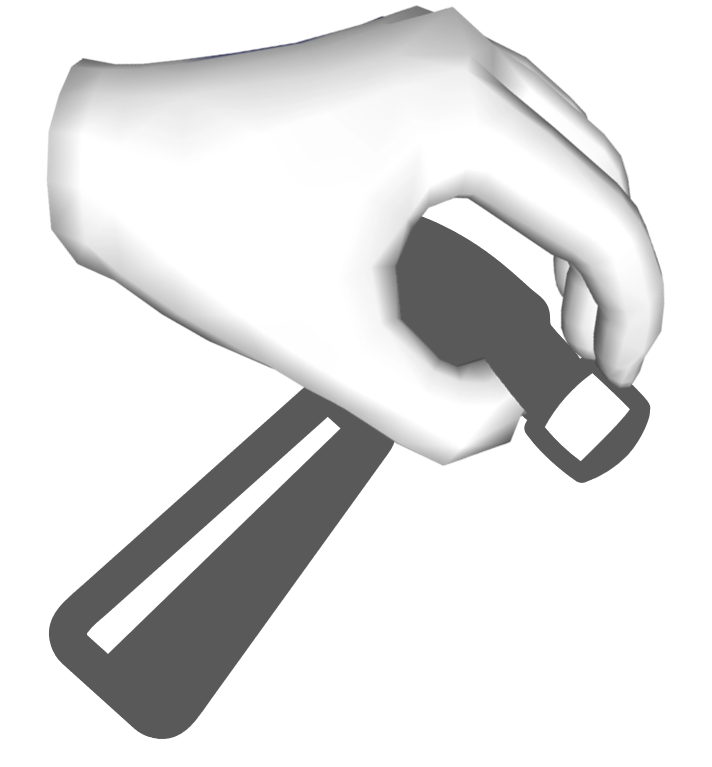}  &\includegraphics[width=0.06\textwidth]{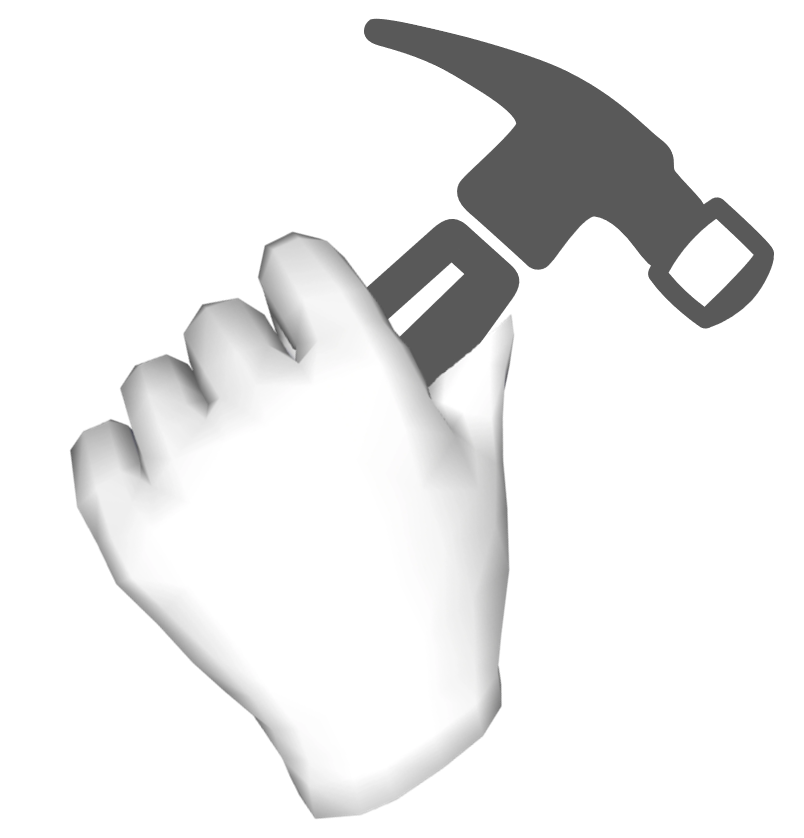}    & \includegraphics[width=0.06\textwidth]{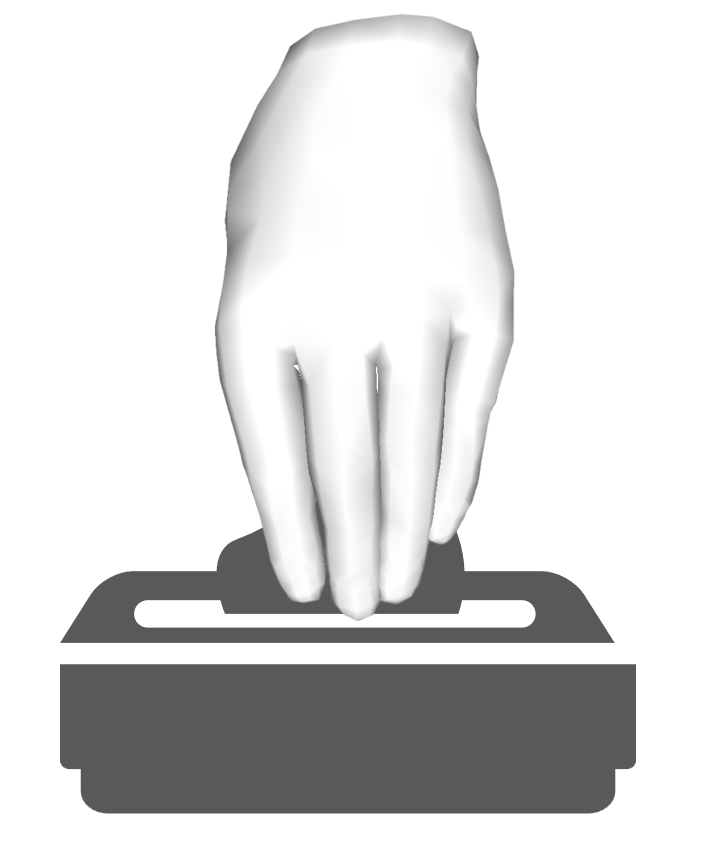} & \\
        \midrule
        GPT-4o & 5/20 & 10/20 & 7/20 & 9/20 & 4/20 & 12/20 &\textbf{ 7/20} & 8/20 & 11/20 & 40.56\% 
        \\
        Qwen-VL & 4/20 & 8/20 & 5/20 & 7/20 & 5/20 & 10/20 & 4/20 & 6/20 & 9/20 & 32.22\% \\
        \midrule
        Robo-ABC & 1/20 & 10/20 & 3/20 & 8/20 & 3/20 & 10/20 & 2/20 & 10/20 & 11/20 & 34.44\% \\

        RAM  & 5/20 & \textbf{12/20} & 6/20 & 8/20 & 4/20 & 9/20 & 5/20 & 11/20 & 12/20 & 40.00\% \\
        \textbf{GAT-Grasp (Ours)} & \textbf{9/20} & \textbf{12/20} & \textbf{10/20} & \textbf{10/20} & \textbf{6/20} & \textbf{14/20} & 6/20 & \textbf{12/20} & \textbf{14/20} & \textbf{51.67\%} \\
        \bottomrule
    \end{tabular}
\end{table*}

\subsection{Baseline Methods}

We compare our method against five baselines: GPT-4o\cite{OpenAI2024}, Qwen-VL\cite{bai2023qwen}, Robo-ABC\cite{juRoboABCAffordanceGeneralization2024}, and RAM\cite{kuangRAMRetrievalBasedAffordance2024a}. GPT-4o and Qwen-VL are leading multimodal models that predict spatial affordances using vision grounding capabilities. Robo-ABC and RAM adopt a retrieve-and-transfer approach: Robo-ABC retrieves affordance memories based on visual similarity, while RAM incorporates both image and language queries for affordance retrieval.

For a fair comparison with vision-language models, we first use an object detection model to crop the target object, replicating the function of our pointing gesture in specifying the graspable region. To better leverage the spatial reasoning capabilities of GPT-4o and Qwen-VL, we follow\cite{yuanRoboPointVisionLanguageModel2024}, labeling the input image with coordinate axes and prompting the model with verb-noun phrases (e.g., “Lift the cup handle” or “Grab the cup body”). The models then output a bounding box (top-left and bottom-right corners) around the target region, from which a grasping point is randomly selected. The grasp pose is then generated using the same grasp generation model as in our method. For Robo-ABC, we use our affordance memory for retrieval due to the lack of a public memory bank. For RAM, we provide task-specific language descriptions and follow its original configuration for grasping experiments.

\subsection{Results and Analysis}

\begin{figure}[t]  
    \centering  
    \includegraphics[width=0.48\textwidth]{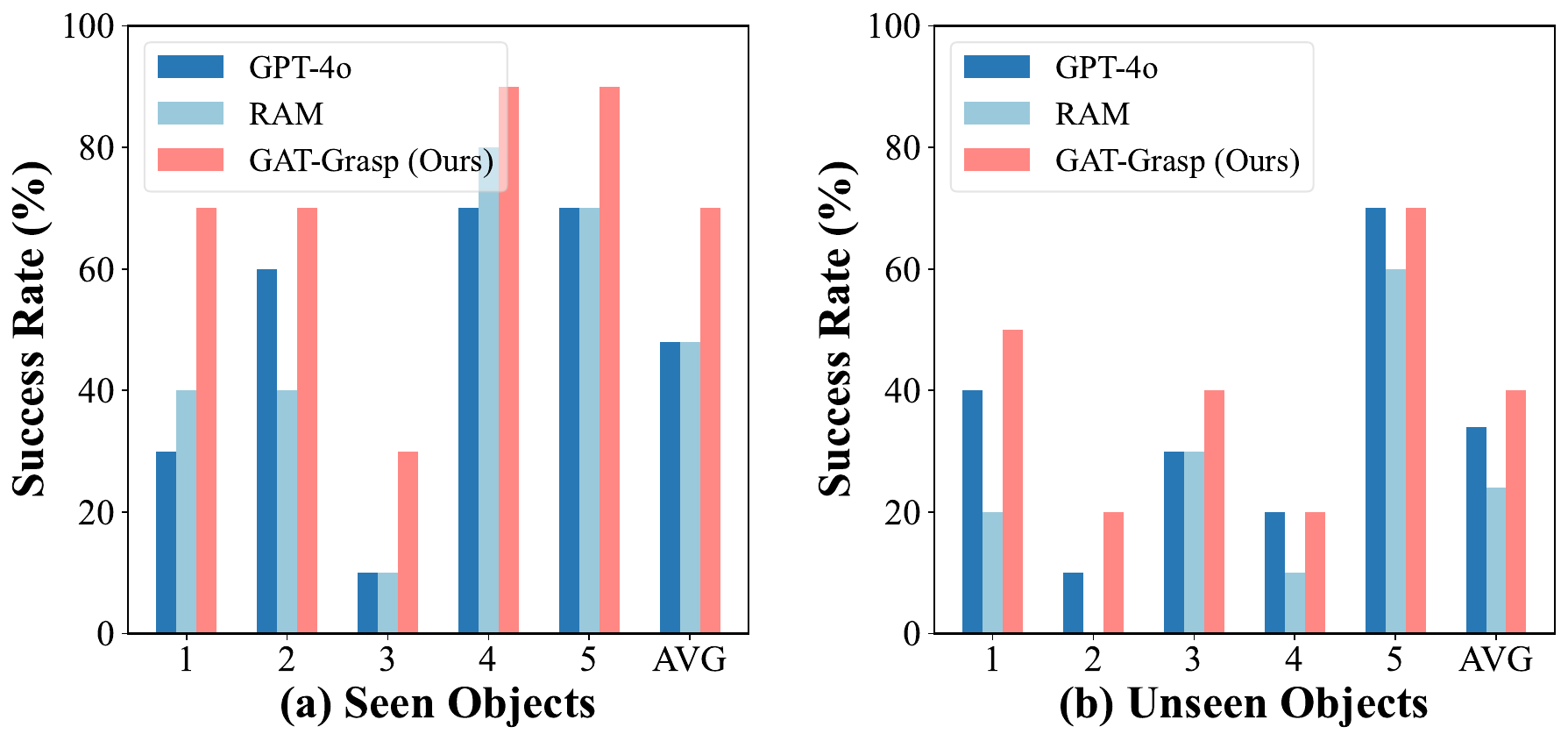}
    \caption{ Quantitative results of single-object grasping experiments. The x-axis represents the object part index: for seen objects, these include the bucket handle, bucket edge, coke ring, coke can body, and stapler; for unseen objects, these include the screwdriver handle, screwdriver tip, teapot handle, teapot lid, and plush toy. Each experiment was repeated 10 times to measure the success rate (SR).  
    }  
    \label{result: single_object}
\end{figure}

Tab. \ref{tab1} presents the comparative results of GAT-Grasp and baseline methods in cluttered scenarios. We use Success Rate (SR) as the primary metric, where a grasp is successful only if the method correctly infers the object’s affordance and executes a stable grasp. GAT-Grasp outperforms all baselines, achieving an average SR of 51.67\%. GPT-4o and Qwen show limitations in spatial reasoning, while Robo-ABC, relying solely on visual similarity, fails to distinguish affordance regions within objects. RAM, despite using both image and language queries, struggles to align language descriptions with spatial affordances.

We attribute the superior performance of GAT-Grasp to three key design choices. First, pointing gestures provide a spatial prior, reducing deviations and improving affordance localization. Second, gestures, as a modality inherently correlated with spatial positions, offer a more effective integration with object geometry than language, enabling precise grasp point prediction. Third, constraints on gripper rotation angles prevent collision-prone grasps, enhancing adaptability in complex scenarios. These features collectively contribute to GAT-Grasp's robustness and high success rate.

We also conduct single-object grasping experiments on both seen and unseen objects, as shown in Fig. \ref{result: single_object}. Here, unseen objects refer to object categories not present in our affordance memory bank. The results show that GAT-Grasp outperforms GPT-4o and RAM on seen objects. More importantly, our method maintains strong performance on unseen categories, demonstrating its ability to generalize affordance reasoning beyond memorized instances.

\subsection{Ablation Studies}

\begin{table}[h]
    \centering
    \renewcommand{\arraystretch}{1.5} 
    \setlength{\tabcolsep}{14pt}  
    \caption{Ablation studies on different components of our method.}
    \label{tab: ablation}
    \begin{tabular}{lcc}
        \toprule
        Ablation Method &SR $\uparrow$ & DTM $\downarrow$ \\
        \midrule
        w/o Pointing Gesture & 13.33\% & 0.291 \\
        w/o Affordance Transfer & 48.89\% & 0.061 \\
        w/o Rotation Mapping & 42.22\%& - \\

        w/o Grasp Gesture &35.56\%  &0.061  \\
        w/o Grasp Generation Model  & 44.44\% & - \\
        \midrule
        \textbf{Full Pipeline} & \textbf{51.67\%} & \textbf{0.052} \\
        \midrule
        SD $\rightarrow$ SD-DINOv2 & 51.11\% & 0.058 \\
        SD $\rightarrow$ CLIP & 22.22\% & 0.092 \\
        \bottomrule
    \end{tabular}
\end{table}

To better understand the contribution of each component in GAT-Grasp, we conduct ablation experiments by selectively removing or modifying key modules: (1) pointing gestures, (2) grasp gestures, (3) affordance retrieval within grasp gestures, (4) hand-gripper mapping, and (5) the grasp generation model, along with alternative semantic correspondence models. All experiments are conducted 5 times in cluttered scenes, with average results reported.

We assess performance using Success Rate (SR) and Distance to Mask (DTM), following the methodology in \cite{juRoboABCAffordanceGeneralization2024}. SR measures the overall grasp success, while DTM quantifies the accuracy of affordance localization by computing the shortest distance between the predicted affordance position and the ground truth mask, normalized by the diagonal length of the image.

As shown in Tab. \ref{tab: ablation}, removing pointing gestures significantly degrades both SR and DTM. The pointing gesture serves as an effective prior, guiding the model to achieve more accurate and stable grasping by reducing ambiguity caused by relying solely on the grasp gesture. Without affordance transfer, even slight deviations in pointing gestures can alter the predicted grasping point, negatively impacting results. Similarly, removing rotation mapping reduces success rates, highlighting the importance of precise gripper angles. Replacing the visual base model also harms performance, as different models vary in spatial and contextual understanding. In our experiments, the SD model achieves the best results.

Notably, removing the grasp generation model (HGGD) has a relatively small impact, demonstrating that our method can operate effectively without relying on it. This flexibility allows GAT-Grasp to function without a mandatory grasp planner, using it only as an optional enhancement. Overall, the ablation experiments validate the rationality and effectiveness of our method, emphasizing the critical roles of pointing gestures, affordance transfer, and a high-quality visual base model in achieving robust grasping performance.

\section{CONCLUSIONS}

In this work, we introduced GAT-Grasp, a novel gesture-driven framework for task-aware robotic grasping that bridges the gap between human intuition and robotic execution. Our method combines pointing gestures to locate graspable regions and grasp gestures to refine grasp positions and orientations, enabling zero-shot affordance transfer without additional training. Real-world experiments demonstrate the robustness of GAT-Grasp in diverse scenarios, showcasing its ability to handle unseen objects and cluttered environments. The framework's extensibility to tasks such as pick-and-place and bimanual manipulation highlights its potential for broader applications in human-robot collaboration and assistive robotics. Looking ahead, we plan to explore more complex gesture-based interactions, enhance adaptability in dynamic environments, and investigate its potential to support individuals with communication impairments through intuitive, non-verbal interaction modalities.







{\bibliographystyle{IEEEtran}
\bibliography{iros2025}}

\begin{thebibliography}{10}
\providecommand{\url}[1]{#1}
\csname url@rmstyle\endcsname
\providecommand{\newblock}{\relax}
\providecommand{\bibinfo}[2]{#2}
\providecommand\BIBentrySTDinterwordspacing{\spaceskip=0pt\relax}
\providecommand\BIBentryALTinterwordstretchfactor{4}
\providecommand\BIBentryALTinterwordspacing{\spaceskip=\fontdimen2\font plus
\BIBentryALTinterwordstretchfactor\fontdimen3\font minus \fontdimen4\font\relax}
\providecommand\BIBforeignlanguage[2]{{%
\expandafter\ifx\csname l@#1\endcsname\relax
\typeout{** WARNING: IEEEtran.bst: No hyphenation pattern has been}%
\typeout{** loaded for the language `#1'. Using the pattern for}%
\typeout{** the default language instead.}%
\else
\language=\csname l@#1\endcsname
\fi
#2}}

\bibitem{kuangRAMRetrievalBasedAffordance2024a}
Y.~Kuang, J.~Ye, H.~Geng, J.~Mao, C.~Deng, L.~Guibas, H.~Wang, and Y.~Wang, ``Ram: Retrieval-based affordance transfer for generalizable zero-shot robotic manipulation,'' in \emph{8th Annual Conference on Robot Learning}, Sept. 2024.

\bibitem{maGLOVERGeneralizableOpenVocabulary2024}
T.~Ma, Z.~Wang, J.~Zhou, M.~Wang, and J.~Liang, ``Glover: Generalizable open-vocabulary affordance reasoning for task-oriented grasping,'' Nov. 2024.

\bibitem{9395580}
K.-B. Park, S.~H. Choi, J.~Y. Lee, Y.~Ghasemi, M.~Mohammed, and H.~Jeong, ``Hands-free human–robot interaction using multimodal gestures and deep learning in wearable mixed reality,'' \emph{IEEE Access}, vol.~9, pp. 55\,448--55\,464, 2021.

\bibitem{tsagkasClickGraspZeroShot2024}
N.~Tsagkas, J.~Rome, S.~Ramamoorthy, O.~M. Aodha, and C.~X. Lu, ``Click to grasp: Zero-shot precise manipulation via visual diffusion descriptors,'' Mar. 2024.

\bibitem{nohGraspSAMWhenSegment2024}
S.~Noh, J.~Kim, D.~Nam, S.~Back, R.~Kang, and K.~Lee, ``Graspsam: When segment anything model meets grasp detection,'' Sept. 2024.

\bibitem{constantin2022interactive}
S.~Constantin, F.~I. Eyiokur, D.~Yaman, L.~B{\"a}rmann, and A.~Waibel, ``Interactive multimodal robot dialog using pointing gesture recognition,'' in \emph{European conference on computer vision}.\hskip 1em plus 0.5em minus 0.4em\relax Springer, 2022, pp. 640--657.

\bibitem{medeiros20213d}
A.~C. Medeiros, P.~Ratsamee, J.~Orlosky, Y.~Uranishi, M.~Higashida, and H.~Takemura, ``3d pointing gestures as target selection tools: guiding monocular uavs during window selection in an outdoor environment,'' \emph{ROBOMECH journal}, vol.~8, pp. 1--19, 2021.

\bibitem{brohan2023can}
A.~Brohan, Y.~Chebotar, C.~Finn, K.~Hausman, A.~Herzog, D.~Ho, J.~Ibarz, A.~Irpan, E.~Jang, R.~Julian, \emph{et~al.}, ``Do as i can, not as i say: Grounding language in robotic affordances,'' in \emph{Conference on robot learning}.\hskip 1em plus 0.5em minus 0.4em\relax PMLR, 2023, pp. 287--318.

\bibitem{luVLGrasp6DofInteractive2023}
Y.~Lu, Y.~Fan, B.~Deng, F.~Liu, Y.~Li, and S.~Wang, ``Vl-grasp: A 6-dof interactive grasp policy for language-oriented objects in cluttered indoor scenes,'' in \emph{2023 IEEE/RSJ International Conference on Intelligent Robots and Systems (IROS)}, Oct. 2023, pp. 976--983.

\bibitem{zhengGaussianGrasper3DLanguage2024a}
Y.~Zheng, X.~Chen, Y.~Zheng, S.~Gu, R.~Yang, B.~Jin, P.~Li, C.~Zhong, Z.~Wang, L.~Liu, C.~Yang, D.~Wang, Z.~Chen, X.~Long, and M.~Wang, ``Gaussiangrasper: 3d language gaussian splatting for open-vocabulary robotic grasping,'' Mar. 2024.

\bibitem{damenScalingEgocentricVision2018}
D.~Damen, H.~Doughty, G.~M. Farinella, S.~Fidler, A.~Furnari, E.~Kazakos, D.~Moltisanti, J.~Munro, T.~Perrett, W.~Price, and M.~Wray, ``Scaling egocentric vision: The epic-kitchens dataset,'' in \emph{Proceedings of the European Conference on Computer Vision (ECCV)}, 2018, pp. 720--736.

\bibitem{grauman2022ego4d}
K.~Grauman, A.~Westbury, E.~Byrne, Z.~Chavis, A.~Furnari, R.~Girdhar, J.~Hamburger, H.~Jiang, M.~Liu, X.~Liu, \emph{et~al.}, ``Ego4d: Around the world in 3,000 hours of egocentric video,'' in \emph{Proceedings of the IEEE/CVF conference on computer vision and pattern recognition}, 2022, pp. 18\,995--19\,012.

\bibitem{liuHOI4D4DEgocentric2022}
Y.~Liu, Y.~Liu, C.~Jiang, K.~Lyu, W.~Wan, H.~Shen, B.~Liang, Z.~Fu, H.~Wang, and L.~Yi, ``Hoi4d: A 4d egocentric dataset for category-level human-object interaction,'' in \emph{Proceedings of the IEEE/CVF Conference on Computer Vision and Pattern Recognition}, 2022, pp. 21\,013--21\,022.

\bibitem{goodrichHumanRobotInteraction2008}
M.~A. Goodrich and A.~C. Schultz, ``Human--robot interaction: A survey,'' \emph{Foundations and Trends{\textregistered} in Human--Computer Interaction}, vol.~1, no.~3, pp. 203--275, Jan. 2008.

\bibitem{bonarini2020communication}
A.~Bonarini, ``Communication in human-robot interaction,'' \emph{Current Robotics Reports}, vol.~1, no.~4, pp. 279--285, 2020.

\bibitem{lu2023vl}
Y.~Lu, Y.~Fan, B.~Deng, F.~Liu, Y.~Li, and S.~Wang, ``Vl-grasp: a 6-dof interactive grasp policy for language-oriented objects in cluttered indoor scenes,'' in \emph{2023 IEEE/RSJ International Conference on Intelligent Robots and Systems (IROS)}.\hskip 1em plus 0.5em minus 0.4em\relax IEEE, 2023, pp. 976--983.

\bibitem{tziafasOpenWorldGraspingLarge2024a}
G.~Tziafas and H.~Kasaei, ``Towards open-world grasping with large vision-language models,'' in \emph{8th Annual Conference on Robot Learning}, Sept. 2024.

\bibitem{jojic2000detection}
N.~Jojic, B.~Brumitt, B.~Meyers, S.~Harris, and T.~Huang, ``Detection and estimation of pointing gestures in dense disparity maps,'' in \emph{Proceedings Fourth IEEE International Conference on Automatic Face and Gesture Recognition (Cat. No. PR00580)}.\hskip 1em plus 0.5em minus 0.4em\relax IEEE, 2000, pp. 468--475.

\bibitem{dhingra2020recognition}
N.~Dhingra, E.~Valli, and A.~Kunz, ``Recognition and localisation of pointing gestures using a rgb-d camera,'' in \emph{International Conference on Human-Computer Interaction}.\hskip 1em plus 0.5em minus 0.4em\relax Springer, 2020, pp. 205--212.

\bibitem{das2021data}
S.~S. Das, ``A data-set and a method for pointing direction estimation from depth images for human-robot interaction and vr applications,'' in \emph{2021 IEEE International Conference on Robotics and Automation (ICRA)}.\hskip 1em plus 0.5em minus 0.4em\relax IEEE, 2021, pp. 11\,485--11\,491.

\bibitem{azari2019commodifying}
B.~Azari, A.~Lim, and R.~Vaughan, ``Commodifying pointing in hri: simple and fast pointing gesture detection from rgb-d images,'' in \emph{2019 16th Conference on Computer and Robot Vision (CRV)}.\hskip 1em plus 0.5em minus 0.4em\relax IEEE, 2019, pp. 174--180.

\bibitem{huang2023voxposer}
W.~Huang, C.~Wang, R.~Zhang, Y.~Li, J.~Wu, and L.~Fei-Fei, ``Voxposer: Composable 3d value maps for robotic manipulation with language models,'' \emph{Proceedings of Machine Learning Research}, vol. 229, 2023.

\bibitem{qianThinkGraspVisionLanguageSystem2024}
Y.~Qian, X.~Zhu, O.~Biza, S.~Jiang, L.~Zhao, H.~Huang, Y.~Qi, and R.~Platt, ``Thinkgrasp: A vision-language system for strategic part grasping in clutter,'' July 2024.

\bibitem{kirillov2023segment}
A.~Kirillov, E.~Mintun, N.~Ravi, H.~Mao, C.~Rolland, L.~Gustafson, T.~Xiao, S.~Whitehead, A.~C. Berg, W.-Y. Lo, \emph{et~al.}, ``Segment anything,'' in \emph{Proceedings of the IEEE/CVF International Conference on Computer Vision}, 2023, pp. 4015--4026.

\bibitem{sunGoingDenserOpenVocabulary2023a}
P.~Sun, S.~Chen, C.~Zhu, F.~Xiao, P.~Luo, S.~Xie, and Z.~Yan, ``Going denser with open-vocabulary part segmentation,'' in \emph{Proceedings of the IEEE/CVF International Conference on Computer Vision}, 2023, pp. 15\,453--15\,465.

\bibitem{yuanRoboPointVisionLanguageModel2024}
W.~Yuan, J.~Duan, V.~Blukis, W.~Pumacay, R.~Krishna, A.~Murali, A.~Mousavian, and D.~Fox, ``Robopoint: A vision-language model for spatial affordance prediction for robotics,'' June 2024.

\bibitem{8793804}
A.~Shafti, P.~Orlov, and A.~A. Faisal, ``Gaze-based, context-aware robotic system for assisted reaching and grasping,'' in \emph{2019 International Conference on Robotics and Automation (ICRA)}, 2019, pp. 863--869.

\bibitem{ryu2019gg}
K.~Ryu, J.-J. Lee, and J.-M. Park, ``Gg interaction: a gaze--grasp pose interaction for 3d virtual object selection,'' \emph{Journal on Multimodal User Interfaces}, vol.~13, no.~4, pp. 383--393, 2019.

\bibitem{wang2023you}
S.~Wang, W.~Zhang, Z.~Zhou, J.~Cao, Z.~Chen, K.~Chen, B.~Li, and Z.~Kan, ``What you see is what you grasp: User-friendly grasping guided by near-eye-tracking,'' in \emph{2023 IEEE International Conference on Development and Learning (ICDL)}.\hskip 1em plus 0.5em minus 0.4em\relax IEEE, 2023, pp. 194--199.

\bibitem{li2024okami}
J.~Li, Y.~Zhu, Y.~Xie, Z.~Jiang, M.~Seo, G.~Pavlakos, and Y.~Zhu, ``Okami: Teaching humanoid robots manipulation skills through single video imitation,'' in \emph{8th Annual Conference on Robot Learning}, 2024.

\bibitem{zhouYouOnlyTeach2025}
H.~Zhou, R.~Wang, Y.~Tai, Y.~Deng, G.~Liu, and K.~Jia, ``You only teach once: Learn one-shot bimanual robotic manipulation from video demonstrations,'' Jan. 2025.

\bibitem{juRoboABCAffordanceGeneralization2024}
Y.~Ju, K.~Hu, G.~Zhang, G.~Zhang, M.~Jiang, and H.~Xu, ``Robo-abc: Affordance generalization beyond categories via semantic correspondence for robot manipulation,'' Jan. 2024.

\bibitem{radford2021learning}
A.~Radford, J.~W. Kim, C.~Hallacy, A.~Ramesh, G.~Goh, S.~Agarwal, G.~Sastry, A.~Askell, P.~Mishkin, J.~Clark, \emph{et~al.}, ``Learning transferable visual models from natural language supervision,'' in \emph{International conference on machine learning}.\hskip 1em plus 0.5em minus 0.4em\relax PMLR, 2021, pp. 8748--8763.

\bibitem{potamiasWiLoREndtoend3D2024}
R.~A. Potamias, J.~Zhang, J.~Deng, and S.~Zafeiriou, ``Wilor: End-to-end 3d hand localization and reconstruction in-the-wild,'' Sept. 2024.

\bibitem{pavlakos2024reconstructing}
G.~Pavlakos, D.~Shan, I.~Radosavovic, A.~Kanazawa, D.~Fouhey, and J.~Malik, ``Reconstructing hands in 3d with transformers,'' in \emph{Proceedings of the IEEE/CVF Conference on Computer Vision and Pattern Recognition}, 2024, pp. 9826--9836.

\bibitem{xu2023iterative}
G.~Xu, X.~Wang, X.~Ding, and X.~Yang, ``Iterative geometry encoding volume for stereo matching,'' in \emph{Proceedings of the IEEE/CVF Conference on Computer Vision and Pattern Recognition}, 2023, pp. 21\,919--21\,928.

\bibitem{fischler1981random}
M.~A. Fischler and R.~C. Bolles, ``Random sample consensus: a paradigm for model fitting with applications to image analysis and automated cartography,'' \emph{Communications of the ACM}, vol.~24, no.~6, pp. 381--395, 1981.

\bibitem{tangEmergentCorrespondenceImage2023}
L.~Tang, M.~Jia, Q.~Wang, C.~P. Phoo, and B.~Hariharan, ``Emergent correspondence from image diffusion,'' \emph{Advances in Neural Information Processing Systems}, vol.~36, pp. 1363--1389, Dec. 2023.

\bibitem{chen2023efficient}
S.~Chen, W.~Tang, P.~Xie, W.~Yang, and G.~Wang, ``Efficient heatmap-guided 6-dof grasp detection in cluttered scenes,'' \emph{IEEE Robotics and Automation Letters}, 2023.

\bibitem{fang2023anygrasp}
H.-S. Fang, C.~Wang, H.~Fang, M.~Gou, J.~Liu, H.~Yan, W.~Liu, Y.~Xie, and C.~Lu, ``Anygrasp: Robust and efficient grasp perception in spatial and temporal domains,'' \emph{IEEE Transactions on Robotics}, vol.~39, no.~5, pp. 3929--3945, 2023.

\bibitem{OpenAI2024}
\BIBentryALTinterwordspacing
OpenAI. (2024, May) Hello gpt-4o. [Online]. Available: \url{https://openai.com/index/hello-gpt-4o}
\BIBentrySTDinterwordspacing

\bibitem{bai2023qwen}
J.~Bai, S.~Bai, S.~Yang, S.~Wang, S.~Tan, P.~Wang, J.~Lin, C.~Zhou, and J.~Zhou, ``Qwen-vl: A frontier large vision-language model with versatile abilities,'' \emph{arXiv preprint arXiv:2308.12966}, vol.~1, no.~2, p.~3, 2023.

\end{thebibliography}

\end{document}